\documentclass{ecai} 
\usepackage{latexsym}
\usepackage{amssymb}
\usepackage{amsmath}
\usepackage{amsthm}
\usepackage{booktabs}
\usepackage{enumitem}
\usepackage{graphicx}
\usepackage{color}

\usepackage{upgreek}
\usepackage{todonotes}
\usepackage{subfigure}
\usepackage{amsfonts}
\usepackage{amsthm}
\usepackage{amssymb}
\usepackage{array}
\newcommand{\xti}[2]{#1_t^{#2}}

\newtheorem{proposition}{Proposition}

\newcommand{\BibTeX}{B\kern-.05em{\sc i\kern-.025em b}\kern-.08em\TeX}

\begin{document}

%%%%%%%%%%%%%%%%%%%%%%%%%%%%%%%%%%%%%%%%%%%%%%%%%%%%%%%%%%%%%%%%%%%%%%%%

\begin{frontmatter}

%%% Use this command to specify your submission number.
%%% In doubleblind mode, it will be printed on the first page.

\paperid{1110} 

%%% Use this command to specify the title of your paper.

\title{Distributed Detection of Adversarial Attacks in Multi-Agent Reinforcement Learning with Continuous Action Space}

%%% Use this combinations of commands to specify all authors of your 
%%% paper. Use \fnms{} and \snm{} to indicate everyone's first names 
%%% and surname. This will help the publisher with indexing the 
%%% proceedings. Please use a reasonable approximation in case your 
%%% name does not neatly split into "first names" and "surname".
%%% Specifying your ORCID digital identifier is optional. 
%%% Use the \thanks{} command to indicate one or more corresponding 
%%% authors and their email address(es). If so desired, you can specify
%%% author contributions using the \footnote{} command.

\author[A]{\fnms{Kiarash}~\snm{Kazari}\thanks{Corresponding Author. Email: kkazari@kth.se.}}
\author[A]{\fnms{Ezzeldin}~\snm{Shereen}}
\author[A]{\fnms{Gy\"{o}rgy}~\snm{D\'{a}n}} 

\address[A]{KTH Royal Institute of Technology, Stockholm, Sweden}
%\address[B]{Short Affiliation of Second Author and Third Author}
%\address[C]{Short Alternate Affiliation of Third Author}

%%% Use this environment to include an abstract of your paper.

\begin{abstract}
 We address the problem of detecting adversarial attacks against cooperative multi-agent reinforcement learning with continuous action space. We propose a decentralized detector that relies solely on the local observations of the agents and makes use of a statistical characterization of the normal behavior of observable agents. The proposed detector utilizes deep neural networks to approximate the normal behavior of agents as parametric multivariate Gaussian distributions. Based on the predicted density functions, we define a normality score and provide a characterization of its mean and variance. This characterization allows us to employ a two-sided CUSUM procedure for detecting deviations of the normality score from its mean, serving as a detector of anomalous behavior in real-time. 
We evaluate our scheme on various multi-agent PettingZoo benchmarks against different state-of-the-art attack methods, and our results demonstrate the effectiveness of our method in detecting impactful adversarial attacks. Particularly, it outperforms the discrete counterpart by achieving AUC-ROC scores of over 0.95 against the most impactful attacks in all evaluated environments.
\end{abstract}

\end{frontmatter}

%%%%%%%%%%%%%%%%%%%%%%%%%%%%%%%%%%%%%%%%%%%%%%%%%%%%%%%%%%%%%%%%%%%%%%%%

\section{Introduction}

In recent years we have witnessed enormous success for multi-agent reinforcement learning (MARL) algorithms, with potential applications in autonomous driving, 5G networks, robotics, and smart grid control~\cite{Canese2021_MARL_Apps2}. Based on sequences of observations and rewards from the environment, cooperative MARL (c-MARL) enables a group of agents to perform complex sequential tasks by learning an optimal distributed policy. Besides enabling distributed execution of tasks, c-MARL has been shown to consistently outperform centralized and single-agent RL approaches, especially for complex tasks~\cite{Canese2021_MARL_Apps2}. %Depending on the environment, MARL problems can be cooperative (c-MARL), in which case the agents receive a common reward (or positively correlated rewards), or competitive, in which case the received rewards are negatively correlated.
 
Despite the apparent success of c-MARL, its adoption in safety-critical applications will be dependent on its resilience to noise, faults, and adversarial manipulations. In particular, motivated by the rise of adversarial evasion attacks in machine learning and computer vision~\cite{Goodfellow2014_advexamples}, the trustworthiness of seemingly high-performing data-driven algorithms is becoming increasingly important. In the context of c-MARL for example, an adversary that could compromise an agent could cause it to take sub-optimal actions (either through direct manipulation, or indirectly through manipulating its observations), so as to minimize the team reward~\cite{lin2020_robustnessCMARL}. 
One approach to address adversarial manipulations in MARL is to improve its resilience, e.g., using robust training through dataset augmentation or adversarial regularization~\cite{bukharin2023robust}. Nonetheless, even though this can mitigate the effect of adversarial manipulations to some extent, it does not provide situational awareness. 

Situational awareness requires the timely detection of adversarial activity, and allows to remove or replace misbehaving agents in real-time, or to adjust the policy of non-compromised agents. 
Existing anomaly detection schemes for single-agent RL ~\cite{sedlmeier2020_entropyAD,Zhang2024tmlr,haider2023_singleRL_ood,wang2024_anomaly_offline} are centralized and primarily focus on self-diagnosis by detecting manipulations in the sequence of observations or states. However, in many MARL applications, the agents' observations are private, hence these approaches {\color{black}are} ineffective, especially against a powerful adversary capable of not only manipulating observations but also the victim's actions. {\color{black}Moreover, applying these approaches in a centralized manner to MARL would be limited to detecting only the presence of an attack, without being able to identify the specific victim.}

Decentralized detection schemes for MARL, on the other hand, assume that the agents' action spaces are discrete and assign probabilities to all actions of the agents~\cite{lischke2022_lstm_based_anomaly,kia2023_decentralized_anomaly}. However, many real-world applications of c-MARL involve agents operating in continuous action spaces. For instance, in robotics, agents often need to control motors or actuators with continuous control signals~\cite{Canese2021_MARL_Apps2}, or need to set voltage and frequency for the control of inverter-based resources in smart grids~\cite{Yan2022_marlsmartgrid}. Adapting discrete-action schemes to continuous action spaces would require one to quantize the action space. {\color{black}Importantly, achieving acceptable detection accuracy would demand many quantization levels per action dimension; otherwise, critical information about the structure and different classes of actions in continuous space can be lost~\cite{zhu2024_discretizingContinuous_TNNLS}. The alternative, assuming independence among different action dimensions is a strong assumption that often does not hold~\cite{dadashi2022_ContinuousQuantization_ICML}. Thus, fine-grained quantization is necessary, but it leads to a complexity growing exponentially with the number of action dimensions.} As a result, in high-dimensional action spaces, which are common in applications such as robotic control~\cite{sakryukin2020_HighDimensional_Robotics}, existing detection methods become effectively impractical. Instead, a low-complexity method is needed specifically designed for MARL with continuous action spaces to detect adversarial attacks effectively. 

In this paper, we propose a decentralized approach for detecting adversarial attacks in {\color{black}model-free}  c-MARL with continuous action space. Our contributions are as follows: 
\begin{enumerate}
    \item We utilize the observations of agents to train neural networks that predict the actions of other agents as parameterized multivariate Gaussian distributions.
    \item {\color{black}We use the predicted distribution for computing a normality score, which quantifies how well individual agents conform to their predicted behavior. We provide an analytical characterization of the mean normality score, which allows us to cast attack detection as a mean shift detection problem, and we propose to use the CUSUM method as a solution.}
    \item Through extensive simulations, we demonstrate the effectiveness of the proposed scheme in detecting state-of-the-art attacks in four benchmark MARL environments with continuous action space. {\color{black} Furthermore, we show that our proposed method outperforms the state-of-the-art discrete-action alternative in terms of performance and computational complexity.}
\end{enumerate}
 
The rest of this paper is organized as follows. Section~\ref{section_related} discusses previous work on adversarial attacks against MARL and countermeasures. Section~\ref{section_sysmodel} presents the system model and formulates the problem of decentralized attack detection. Section~\ref{sec:proposed_method} presents the proposed detection approach. The performance of the proposed method is evaluated in Section~\ref{section_eval}.  Section~\ref{section_conc} concludes the paper.

\section{Related Work}
\label{section_related}
Many studies addressing adversarial attacks in single-agent reinforcement learning focus on perturbing states or observations. In these studies, the adversary manipulates observations so that the agent makes sub-optimal decisions with respect to the actual state of the environment~\cite{pattanaik2017_robustDRLwithAdv},\cite{behzadan2017_vulnerability}, \cite{russo2021_towardsOptAtt}, \cite{mo2022_DRL_Attack_TDSC}.  
%In a related study, \cite{mo2022_DRL_Attack_TDSC} addressed the challenge of adversarial attacks on observations with a limited number of attack injections, proposing a decoupled adversarial policy. The policy comprises two sub-policies: one to decide whether to execute an attack injection and another to select the adversarial action in the event of an attack injection. 
In the context of RL with continuous action spaces, \cite{tekgul2022_real_time} proposed real-time adversarial state manipulation attacks against Deep RL and showed the effectiveness of such attacks in MuJoCo~\cite{mujoco} 
physics-based tasks like Walker2d.  

In the domain of MARL,~\cite{lin2020_robustnessCMARL} considered perturbing the observations of a single agent in a c-MARL system. In~\cite{elhami2022_cluster_MARL}, the authors proposed state manipulation adversarial attacks on cluster-based, heterogeneous MARL systems. More powerful attack models in the literature attack directly the actions of the agents. In the competitive MARL setting,~\cite{gleave2019_adversarialAttackAct} showed that an adversary could deceive an agent by manipulating the actions of the competing agent.~\cite{guo2022_comprehensiveTesting} explored the robustness of state-of-the-art c-MARL algorithms against attacks on observations and actions. In a closely related work, \cite{Figura2021_consensusMARL} studied the effect of adversarial attacks on consensus-based networked MARL algorithms. In a different application domain, \cite{ezzeldin2023_MG_secondary} investigated the vulnerability of MARL-based microgrid voltage control to adversarial attacks against sensor readings. Finally,~\cite{yu2024_robust_commMARL} considered robustness against attacks to messages in communicative MARL.

Related to ours are previous works on anomaly detection for sequential data~\cite{oh2019_sequentialAD_iRL}, \cite{malhotra2016_lstmAD_sequential},\cite{wang2021_OneClassRNN}. However, anomaly detection in the context of RL has received less attention compared to domains like video, audio, and text. Among the works addressing anomaly detection in RL,~\cite{Zhang2024tmlr} introduced a framework for detecting anomalous state observations based on a Gaussian approximation of the state representation space. \cite{sedlmeier2020_entropyAD} proposed an entropy-based anomaly detector for identifying out-of-distribution observations, though not within an adversarial setting. Adversarial detection, i.e., identifying the adversarial samples generated in the state space of deep RL is the focus of \cite{lin2017_detection_RL_vision}. Notably, none of these works addressed anomaly detection in a multi-agent setting. Applying these single-agent schemes to MARL would either necessitate centralized tracking of all agents or implementing anomaly detection locally for each agent. The former may not be feasible in decentralized multi-agent systems due to the resource-intensive nature of collecting data from all agents. The latter is sub-optimal as it does not account for adversaries with full control over the actions of the victim agent. In contrast, our approach involves other agents interacting with the environment for anomaly detection. Regardless of whether the misbehavior of the victim agent results from perturbations in its observations or actions, our method can detect anomalies. 

Concerning multi-agent systems, existing works~\cite{Shames2011_Control,Ye2019_Control} have proposed decentralized defenses against adversarial attacks in control systems. However, these approaches rely on a known model of the system, which is typically not available in MARL applications. Closest to our work are recent works addressing anomaly detection in c-MARL~\cite{kia2023_decentralized_anomaly,lischke2022_lstm_based_anomaly}. These works rely on a categorical characterization of the actions taken by agents, making them applicable only to MARL problems with discrete action space. On the contrary, in this work we propose a detector that is applicable to MARL problems with continuous action space.

\section{System Model and Problem Formulation}
\label{section_sysmodel}
\subsection{System Model}
We consider a cooperative multi-agent environment modeled by a decentralized partially-observable Markov decision process (Dec-POMDP). The Dec-POMDP can be described by a tuple $M=(\mathcal{K},\mathcal{S},\lbrace\mathcal{A}^i\rbrace_{i \in \mathcal{K}}, R, P, \lbrace\mathcal{O}^i\rbrace_{i \in \mathcal{K}}, \gamma)$, where $\mathcal{K}=\lbrace1,2,...,K\rbrace$ is the set of agents, $\mathcal{S}$ is the state space, and $\mathcal{O}^i$ is the set of observations of agent $i$. We let $\mathcal{A}^i\subseteq \mathbb{R}^{d_i}$ be the continuous action space of agent $i$. Moreover, $R$ and $P$, denote the reward function, and the state transition probability, respectively, and $\gamma\in (0,1)$ is a discount factor. At each time step $t\geq0$, agent $i$ observes $\xti{o}{i}\in \mathcal{O}^i$ from the environment and chooses an action $\xti{a}{i}\in \mathcal{A}^i$. The joint action profile $\mathbf{a}_t=\lbrace a_t^i\rbrace$ causes the state of the system to change from $s_t=s$ to $s_{t+1}=s'$ with probability $P(s'|s,\mathbf{a}_t)$, and a shared reward $R_t=R(s_t,\mathbf{a}_t)$ is received by all agents. In the reinforcement learning setting, which is the focus of this work, it is assumed that $P$ and $R$ are unknown functions. The objective of the agents is to maximize the total discounted average reward, i.e., $\mathbb{E}[\sum_{t=0}^{\infty}\gamma^{t}R_t]$. 

{\color{black}We say that agent $j$ is an \emph{observable neighbor} of agent $i$ if agent $i$ has access to the sequence of actions played by agent $j$. Note that this does not necessarily mean that those actions are in $o_t^i$. Instead, in many real-world applications, such as multi-UAV navigation or autonomous driving, where other agents' positions are part of an agent's observation, their past actions can be inferred (i.e., they are implicitly observable). We denote by $\mathcal{K}^i$ the set of all observable neighbors of agent $i$. We assume that $\mathcal{K}^i$ is constant over time, and that for each agent $j$ there exists at least one agent $i$ such that $j \in \mathcal{K}^i$, which is a prerequisite for distributed detection.} 

We consider that all agents have been trained using a MARL algorithm, and in the deployment time, agent $i$ acts based on a local policy $\pi^i(\tau_t^i)$, where $\tau_t^i \in \Gamma^i\triangleq(\mathcal{O}^i\times\mathcal{A}^i)^*$ is the history of action-observations of agent $i$ up to time $t$, i.e., $\tau_t^i=(o_0^i, a_0^i,..., o_t^i)$. $\pi^i(\tau_t^i)$ can be a stochastic or a deterministic policy.

\subsection{Attack Model}
\begin{figure}[t]
    \centering
    \includegraphics[width=0.7\linewidth]{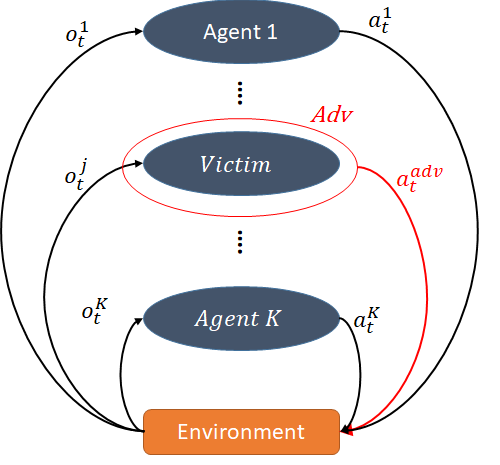}
    \caption{System model with agent $j$ as victim.}
    \label{fig:system_model}
\end{figure}

We consider an adversary that can eavesdrop on the observations and can manipulate the actions of one agent $v\in\mathcal{K}$ in the deployment time of c-MARL. That is, at each time step $t$, the adversary can observe $\xti{o}{adv}=\xti{o}{v}$ from the environment, and can cause the victim agent to take an action $\xti{a}{adv}\in \mathcal{A}^v$ instead of $\xti{a}{v}$, where $\xti{a}{adv}$ follows the adversary's policy $\pi^{adv} \neq \pi^v(\tau_t^v)$. We refer to agent $v$ as the victim agent. This attack model is in line with the models considered in \cite{guo2022_comprehensiveTesting} and \cite{kia2023_decentralized_anomaly}, and is illustrated in Figure~\ref{fig:system_model}. Note that although we assume that only one agent is compromised, our proposed approach is applicable in scenarios involving multiple victims (c.f. Section~\ref{sec::multivictim}).

\subsection{Problem Formulation}
\label{section_problem_formulation}
Consider that the adversary starts to attack some agent $v$ at time step $t_0$ ($t_0$ might be considered $\infty$ in case that no attack happens). Given the sequence of actions of all agents, our objective is to identify the victim agent as soon as possible after the attack starts in a distributed manner, i.e., detection should be done by the agents based on locally available information.  
Consequently, treating  agent $i$ as an observer and agent $j\in \mathcal{K}^i$ as a prospective victim, the considered anomaly detection problem can be formulated as follows. 

\noindent \textbf{Decentralized c-MARL Attack Detection}: Given a history $\tau_{t}^{ij}=o_0^i,a_0^j,o_1^i,a_1^j,...,o_t^i,a_t^j$ at time $t$, find a function $M^{ij}:(\mathcal{O}^i\times\mathcal{A}^j)^* \rightarrow \lbrace 0,1 \rbrace$, s.t. $M^{ij}(\tau_t^{ij})=0$ if the sequence $\tau_{t}^{ij}$ is normal (i.e., it is the consequence of following $\pi^j$ by agent $j$) and $M^{ij}(\tau_t^{ij})=1$ if $\tau_{t}^{ij}$ is abnormal (agent $j$ is a victim). 

%\vspace{-2mm}
\section{Decentralized Attack Detection}
%\vspace{-1mm}
\label{sec:proposed_method}
{\color{black}In this section, we describe our proposed detector, i.e., the algorithm for implementing function $M^{ij}$ for agent $i\in\mathcal{K}$ and agent $j\in\mathcal{K}^i$.} The proposed algorithm is composed of two steps. First, it estimates the distribution of agent $j$'s action (for the upcoming time-step) based on the observations of agent $i$. In the second step, a normality score is computed based on the estimated distribution and the action actually taken by agent $j$; the normality score is then used for detection, relying on a characterization of its mean and standard deviation. Next, we present a detailed description of these two steps. %{\color{black}In the sequel, agents $i$ and $j$ refer to any two agents such that $j\in\mathcal{K}^i$.}

\subsection{Learning Action Distributions}
We propose to  approximate the action distribution $f^{ij}(.|\tau_t^i)$ of agent $j$ from agent $i$'s point of view by a $d_j$-dimensional Gaussian distribution (recall that $d_j$ is the dimension of the action space) parameterized by its mean, $\mu_t^{ij}(\tau_t^i)\in \mathbb{R}^{d_j}$, and its covariance matrix, $\boldsymbol{\Sigma}_t^{ij}(\tau_t^i)\in \mathbb{R}^{d_j\times d_j}$. Observe that the parameters of the distribution are a function of the history $\tau_t^i$. The proposed approximation is motivated by that many deep RL algorithms for continuous action spaces represent a policy as a multivariate Gaussian-distribution~\cite{yang2021_offpcc_RL_algorithms}. While this does not imply that the actions of other agents based on the observation of a single agent would be Gaussian, our results in Section~\ref{section_eval} show that the Gaussian distribution  can effectively approximate the other agents' behaviour based on local observations. The approximation can thus be formulated as
\begin{align}
\label{eq_gaussian_density}
    &f^{ij}(\mathbf{x}|\tau_t^i)\approx \nonumber \\
    &(2\pi)^{-\frac{d_j}{2}}\det(\boldsymbol{\Sigma}_t^{ij})^{-\frac{1}{2}}\text{exp}\left[ -\frac{1}{2}(\mathbf{x}-\mu_t^{ij})^\top (\boldsymbol{\Sigma}_t^{ij})^{-1} (\mathbf{x}-\mu_t^{ij})\right] \nonumber\\
    &\;\;\;\;\;\;\;\;= g(\mathbf{x}; \mu_t^{ij}, \boldsymbol{\Sigma}_t^{ij}).
\end{align}
%We denote the right side of (\ref{eq_gaussian_density}) by $g(\mathbf{x}; \mu_t^{ij}, \boldsymbol{\Sigma}_t^{ij})$. 

The objective of agent $i$ is to learn a function that takes $\tau_t^i$ as the input, and outputs  $\mu_t^{ij}$ and $\boldsymbol{\Sigma}_t^{ij}$. We propose to approximate this function using a recurrent neural network (RNN), represented as $\text{NET}^{ij}$. Our approach is based on the intuition that the task of predicting the parameters of the Gaussian distribution  can be regarded as a regression problem where (the sequence of) agent $i$'s observations are the inputs. In the training phase of $\text{NET}^{ij}$, the objective is to find weights $\boldsymbol{\theta}^{ij}$ such that the outputs, i.e., $\boldsymbol{\mu}$ and $\boldsymbol{\Sigma}$ would  maximize the log-likelihood of observing $a_t^j$ given $\tau_t^i$. Accordingly, $\text{NET}^{ij}$ is trained to maximize the expected log-likelihood $\mathbb{E}[\log f^{ij}(a_t^j|\tau_t^i)]$. Using (\ref{eq_gaussian_density}), the objective function can be expanded as

\begin{align}
\label{eq:loss_func_original}
    &\max_{\boldsymbol{\theta}^{ij}} \mathbb{E}[\log f^{ij}(a_t^j|\tau_t^i)] = \max_{\boldsymbol{\theta}^{ij}} \mathbb{E}[\log g(a_t^j; \mu_t^{ij}, \boldsymbol{\Sigma}_t^{ij})] \nonumber \\
    & =\max_{\boldsymbol{\theta}^{ij}} \mathbb{E}\bigg[- \frac{1}{2} \left(\log \det(\boldsymbol{\Sigma}_t^{ij}) \right. \nonumber\\
    &\qquad \qquad \quad +\left.(a_t^j-\mu_t^{ij})^\top (\boldsymbol{\Sigma}_t^{ij})^{-1} (a_t^j-\mu_t^{ij})\right)\bigg]
\end{align}
Note that $\boldsymbol{\Sigma}_t^{ij}$ is a covariance matrix, hence it should be symmetric and positive definite. In order to ensure this condition is met, we consider the Cholesky factorization \cite{horn2012_matrixBook} of the covariance matrix as $\boldsymbol{\Sigma}_t^{ij}=\mathbf{L}_t^{ij}\mathbf{L}_t^{ij\top}$ where $\mathbf{L}_t^{ij}$ is a lower triangular matrix. Consequently, we treat the non-zero elements of $\mathbf{L}_t^{ij}$ as the outputs of $\text{NET}^{ij}$, which are unconstrained. Now, the objective function in (\ref{eq:loss_func_original}) is equivalent to 
\begin{align}
    &\min_{\boldsymbol{\theta}^{ij}} \mathbb{E}\bigg[  2\log \det(\mathbf{L}_t^{ij}) \nonumber\\
    &\qquad \qquad \quad+ (a_t^j-\mu_t^{ij})^\top (\mathbf{L}_t^{ij}\mathbf{L}_t^{ij\top})^{-1} (a_t^j-\mu_t^{ij})\bigg] \\
    & \label{eq_loss function}\qquad  \quad =\min_{\boldsymbol{\theta}^{ij}} \mathbb{E}\bigg[  2\sum_{k=1}^{d_j}\log l_t^{ij}[k] + y_t^{ij\top}y_t^{ij}\bigg] 
\end{align}
where $l_t^{ij}[k]$ denotes the $k$-th diagonal element of $\mathbf{L}_t^{ij}$ and $y_t^{ij}\triangleq \mathbf{L}_t^{ij^{-1}}(a_t^j-\mu_t^{ij})$. Note that since $\mathbf{L}_t^{ij}$ is a lower triangular matrix, there is no need for inverting it and $y_t^{ij}$ can be computed directly using low complexity algorithms like forward substitution \cite{horn2012_matrixBook}. Training $\text{NET}^{ij}$ is done by collecting samples of the form $(\tau_t^i, a_t^j)$ during the normal execution of the policies $\lbrace\pi^j\rbrace_{j\in\mathcal{K}}$ of the agents, and then optimizing the network's parameters according to the objective   (\ref{eq_loss function}).
\begin{figure}[t]
    \centering
    \includegraphics[width=0.95\linewidth]{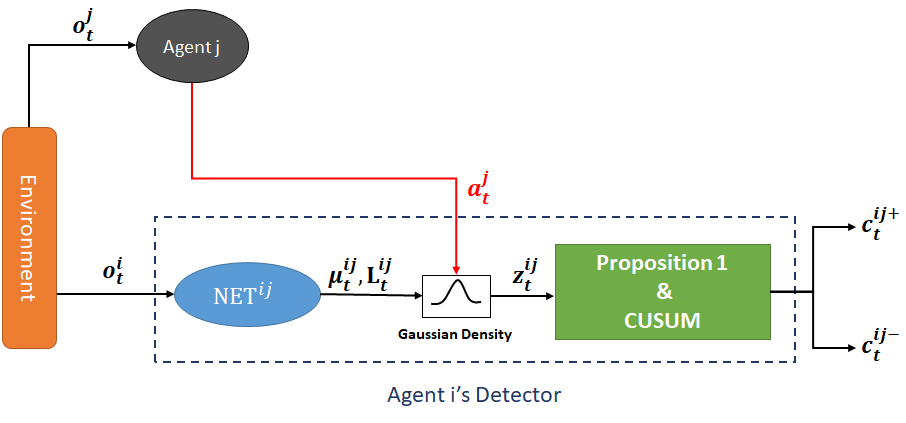}
    \caption{Proposed detection scheme for agent $i$ as the observer and agent $j$ as the potential victim.}
    \label{fig:Detectors}
\end{figure}
\subsection{\color{black}{Normality Score and Detection Procedure}}
We use the action density functions learned in  the previous step for quantifying the normality of the sequence of agent $j$'s actions up to time step $t$. While the problem seemingly resembles that of sequential hypothesis testing~ \cite{lehmann2005_GoodnessofFit}, where the objective is to determine whether a sequence of samples are generated from a given distribution, in our case the distribution is state-dependent and changes over time. Hence, existing approaches to sequential hypothesis testing cannot be applied directly. 

\color{black}
Instead, motivated by the log-likelihood ratio computed in sequential hypothesis testing approaches, we propose to use the normalized log-likelihood based on the approximated action distributions to compute a normality score defined as  
\begin{equation}
\label{eq_normality score}
z_t^{ij}=\log(\frac{f^{ij}(a_t^j|\tau_t^i)}{\max_{\mathbf{x}} f^{ij}(\mathbf{x}|\tau_t^{i})}), 
\end{equation}
which represents the log-likelihood of the predicted density at $a^j_t$ normalized by the maximum value of the density function. Intuitively, the reason for the normalization is to take the confidence of the predictions into account. 

Importantly, if $f^{ij}$ is Gaussian then $z_t^{ij}$ is closely related to the Mahalanobis distance of $a_t^j$ and $f^{ij}(.|\tau_t^{i})$. To see why, observe that $f^{ij}$ attains its maximum at $\mathbf{x}=\mu_t^{ij}$. Thus, 
\begin{align}
     &\qquad\qquad z_t^{ij} = \log\bigg[\frac{g(a^j_t; \mu_t^{ij}, \boldsymbol{\Sigma}_t^{ij})}{g(\mu_t^{ij}; \mu_t^{ij}, \boldsymbol{\Sigma}_t^{ij})}\bigg]  \\ 
    &= -\frac{1}{2}(a^j_t-\mu_t^{ij})^\top (\boldsymbol{\Sigma}_t^{ij})^{-1} (a^j_t-\mu_t^{ij})=-\frac{D(a^j_t, f^{ij}_t)}{2},\nonumber
\end{align}
where $D( .,. )$ denotes the Mahalanobis distance. An essential property of the score we propose for detection is the invariance of its expected value over time. 
\begin{proposition}
\label{th:z_expectation}
    Consider agents $i,j\in\mathcal{K}$ at time $t$, where $j\in\mathcal{K}^i$, and assume $a_t^j|\tau_t^i\sim \mathcal{N}(\mu_t^{ij},\boldsymbol{\Sigma}_t^{ij})$. Then the first and second moments of $z_t^{ij}$ are 
    \begin{equation}
        m_z^j\triangleq\mathbb{E}[z_t^{ij}]=-d^j/2, \quad \sigma_z^{j^2}\triangleq\mathbb{E}[(z_t^{ij}-m_z^j)^2]=d^{j}/2.
    \end{equation}
    
\end{proposition}
\begin{proof}
    For ease of notation, we drop the subscript $t$ and superscripts $i,j$. We can write
\begin{align}
    &\mathbb{E}[z]= -\frac{1}{2}\mathbb{E}[(a-\mu)^\top \Sigma^{-1}(a-\mu)] \nonumber\\
    &=-\frac{1}{2}\mathbb{E}[\text{tr}((a-\mu)^\top \Sigma^{-1}(a-\mu))] \nonumber\\
    &= -\frac{1}{2}\mathbb{E}[\text{tr}( \Sigma^{-1}(a-\mu)^\top(a-\mu))]\nonumber\\
    &= -\frac{1}{2}\text{tr}( \Sigma^{-1}\mathbb{E}[(a-\mu)^\top(a-\mu)])\nonumber\\
    &= -\frac{1}{2}\text{tr}( \Sigma^{-1}\Sigma)=-d/2.
\end{align}
\noindent For the variance, we can write
\begin{equation}
    \mathbb{E}[z^2]= \frac{1}{4}\mathbb{E}[((a-\mu)^\top \Sigma^{-1}(a-\mu))^2]. 
\end{equation}
Since $\Sigma$ is a positive definite matrix, it has a unique symmetric positive semi definite square root $R$, such that $RR=\Sigma$. Then for $x=R^{-1}(a-\mu)$ it holds that $x\sim\mathcal{N}(0,I)$. Thus, we have
\begin{align}
    &\mathbb{E}[z^2]=\frac{1}{4}\mathbb{E}[(x^\top R R^{-1} R^{-1}Rx )^2]=\frac{1}{4}\mathbb{E}[(x^\top x )^2]\nonumber\\
    &=\frac{1}{4}(\text{tr}(I)^2 + 2\text{tr}(I))=\frac{1}{4}(d^2 + 2d).
\end{align}
Accordingly, $\sigma_z^2=\mathbb{E}[z^2]-\mathbb{E}[z]^2=d/2$.
\end{proof} 
Proposition \ref{th:z_expectation} shows that, although the distribution of predicted actions varies over time, as long as the actions of agent $j$ follow the conditional distribution expected by observer agent $i$, the expected value of the normality score remains constant. 

{\color{black}Importantly, Proposition \ref{th:z_expectation} allows us to cast attack detection as the problem of detecting a shift in the mean of a sequence of random variables, for which CUSUM is a well-known stopping rule~\cite{page_cusum}, which is optimal under the assumption that the samples are i.i.d \cite{kazari2025_quickest, xie2021_cusum_survey}. In our setting, the samples $z_t^{ij}$ are not independent; however, the complexity of their dependencies makes it infeasible to derive an optimal stopping rule. We thus propose to apply CUSUM on the normality score $z_t^{ij}$ combined with Proposition \ref{th:z_expectation} for detection. As shown in Section \ref{section_eval}, it performs well despite samples not being independent.} Accordingly, our proposed detection method is based on computing 
\begin{align}
   c_t^{ij^+}&=\max\lbrace 0, c_{t-1}^{ij^+}+ \frac{z_t^{ij}-m_z^j}{\sigma_z^j}-w\rbrace,  \\
   c_t^{ij^-}&=\max\lbrace 0, c_{t-1}^{ij^-}+ \frac{m_z^j - z_t^{ij}}{\sigma_z^j}-w\rbrace, 
\end{align}
which are used to determine a deviation from the mean in the positive and the negative directions, receptively, $w$ is a tunable hyperparameter that determines CUSUM's sensitivity, and $c_{-1}^{ij^+}=c_{-1}^{ij^-}=0$. For detection, each agent $i$ continuously updates $c_t^{ij^+}$ and $ c_t^{ij^-}$ corresponding to every agent $j$ in $\mathcal{K}^i$, and considers agent $j$ as compromised if either $c_t^{ij^+}>\beta^{ij^+}$ or $c_t^{ij^-}>\beta^{ij^-}$, where $\beta^{ij^+}$ and $\beta^{ij^-}$ are predefined thresholds.  We refer to the proposed detection scheme as the Parameterized Gaussian CUSUM (PGC) detector. Figure~\ref{fig:Detectors} shows the overall structure of the proposed detector.

\color{black}
%To simplify the computation of the normality score, we can use \eqref{eq_gaussian_density} and observe that the density function attains its maximum at $\mathbf{x}=\mu_t^{ij}$, $z_t^{ij}$, we thus obtain  
%\begin{align}
%    &\qquad\qquad z_t^{ij} = \log\bigg[\frac{g(\mathbf{x}; \mu_t^{ij}, \boldsymbol{\Sigma}_t^{ij})}{g(\mu; \mu_t^{ij}, \boldsymbol{\Sigma}_t^{ij})}\bigg] \nonumber \\ 
%    &= -\frac{1}{2}(\mathbf{x}-\mu_t^{ij})^\top (\boldsymbol{\Sigma}_t^{ij})^{-1} (\mathbf{x}-\mu_t^{ij})=  -\frac{1}{2}y_t^{ij\top}y_t^{ij}
%\end{align}

The proposed detector can be combined with different decision rules for distributed detection of an attack. A straightforward rule would be to consider an agent as compromised once any of the other agents detects it as attacked. Alternatively, one could consider an agent compromised when at least $u>1$ agents detect that it is compromised.

\section{Evaluation}
\label{section_eval}
We present results obtained using the proposed detector against four types of attacks in four MARL environments. 

\subsection{MARL Environments}
\color{black}
We used four PettingZoo c-MARL continuous action environments~\cite{pettingzoo} for the evaluation, namely, \emph{Multiwalker}~\cite{gupta2017_SISL}, \emph{Tag} (Simple Tag)~\cite{lowe2017_MPE}, \emph{World Comm}~\cite{lowe2017_MPE}, and \emph{Pistonball}. The main properties of the considered scenarios are summarized in Table \ref{tab:envs}. {\color{black} Based on the observations provided by the environments, in Pistonball and Multiwalker, $\mathcal{K}^i$ is the set of the agents adjacent to agent $i$, while in the other two environments, $\mathcal{K}^i$ includes all agents other than $i$.} %For more details we refer to the Appendix. 

\subsection{Attack Strategies}
\label{section_attacks}
We use four attack strategies for the evaluation.
\paragraph{Random Attack (RAND):} 
A naive adversary that at each step chooses an action uniformly at random in $\mathcal{A}^v$.
\paragraph{Action Attack (ACT):} The adversary's objective is to minimize the team reward. The policy of the adversary is obtained by solving a single-agent reinforcement learning problem with the action space defined as $\mathcal{A}^v$ and the objective of minimizing $\mathbb{E}[\sum_{t=0}^{\infty}\gamma^tR_t]$~\cite{guo2022_comprehensiveTesting}. Since the policies of the non-victim agents are fixed, they can be considered part of the environment.
\color{black}
\paragraph{Projected Gradient-Based Attack (GRAD):}
This approach is commonly used in single agent RL to manipulate agent's observations \cite{pattanaik2017_robustDRLwithAdv}. It is based on finding a suitable $l_\infty$-bounded manipulation through a perturbation in the gradient-ascent direction of a loss function, e.g., as in FGSM~\cite{Goodfellow2014_advexamples}. Due to continuity of the action space, we apply this method directly to the actions. Accordingly, we considered the following manipulation strategy based on the Q function:
\begin{equation}
    a_t^{adv}= Proj \lbrace a_t^v - \epsilon sign(\nabla_a Q^v(\tau_t^v, a_t^v))\rbrace,
\end{equation}
where $Q^v(\tau, a)$ is the victim's Q function, and $Proj$ denotes the projection onto $\mathcal{A}^v$.
\color{black}
\begin{table}[t]
        \caption{Number of agents, observation size, action space dimension, and time limit in the considered scenarios. In Tag and World Comm one of the agents does not belong to the c-MARL team.}
    \centering
    \begin{tabular}{ |m{0.22\linewidth}|m{0.09\linewidth}| m{0.22\linewidth}| m{0.1\linewidth}| m{0.1\linewidth}|}
    \hline
        \textbf{Environment} & { Agents} & Obs. Size & Action Dim. & Time Limit\\ \hline
         {Multiwalker}    & 5     &    31 &  4   & 200   \\ \hline
         {Tag}           &  4    &   16  &  5  & 25  \\ \hline
         {WorldComm}   & 5     &    25 &  9 & 25 \\ \hline
         {Pistonball}   & 10     &    RGB (84,84,3)  &  1 & 125\\ \hline         
    \end{tabular}
     \label{tab:envs}
\end{table}

\begin{figure}[t]
    \centering
    \includegraphics[width=\linewidth]{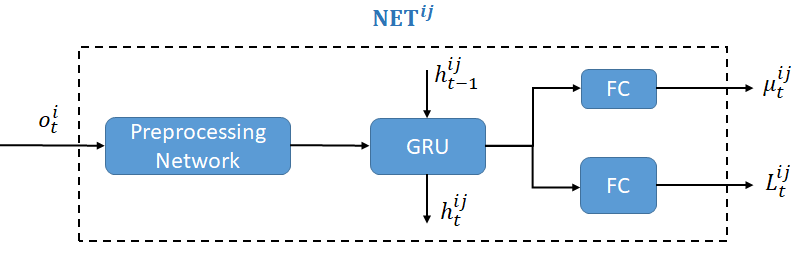}
    \caption{Structure of $\text{NET}^{ij}$ used for prediction.}
    \label{fig:NET}
\end{figure}
\paragraph{Dynamic Adversary (DYN):} The dynamic adversary is a powerful adversary that is aware of the detection scheme and can observe the normality scores that the agents compute. It uses this information for learning attacks that are hard to detect using the proposed detector, as done in~\cite{kia2023_decentralized_anomaly}. Following the notion of an expectedly undetectable attack introduced in ~\cite{kia2023_decentralized_anomaly}, the adversary's problem can be formulated as a constrained RL problem whose solution would be a Markovian policy. 
{\color{black}To tailor this attack against our detection procedure, we consider the reward function for this single agent RL as $R_t^{adv}= -(R_t + \sum_{i: v\in\mathcal{K}^i}\lambda_i |z_t^{iv}-m_z^v|)$, where  $\lambda_i$ is a weight parameter that captures the importance of remaining undetected by agent $i$, and effectively controls the trade-off between detectability and the impact of the attack on the team reward.} 

For instance, for  $\lambda_i=0$ the attack is equivalent to ACT, maximizing attack impact. Conversely, for high values of $\lambda_i$, the objective is dominated by trying to remain undetected by keeping the normality score close to its expectation. For simplicity we consider uniform weights, i.e.,  $\lambda_i=\lambda$. 
 
\color{black}
\subsection{Baselines}
Existing single agent RL anomaly detection methods typically detect anomalies in the observations or states of an agent~\cite{sedlmeier2020_entropyAD,Zhang2024tmlr}. Such detection schemes could be used by the victim, but if used by non-victim agents, they would not be able to identify the victim agent, only the fact that there may be an attack. To provide a fair comparison, we thus focus on baselines that allow to identify victim agents. {\color{black} We provide a comparison to \cite{Zhang2024tmlr} in Appendix C.4.}

As baselines we thus use the detection method proposed in~\cite{kia2023_decentralized_anomaly} for MARL with discrete action spaces, which we refer to as \emph{Discrete}, and a number of variants of our detection method. To adapt \emph{Discrete} to a continuous action space, we uniformly quantize each dimension of the action space into $Q$ intervals, resulting in an action set of size $Q^d$ if the continuous action set has dimension $d$. 

We consider three variants of our detection method: using the Dirichlet distribution, the Beta distribution, and Gaussian with independent components (i.e., a diagonal covariance) as the distribution approximating $f^{ij}$. These variants are referred to as \emph{Dirichlet}, \emph{Beta}, and \emph{Independent PGC (I-PGC)} in this section. %Note that, to use the Beta or Dirichlet distribution, we first apply a linear transformation to map each component of the action vector to the interval $[0,1]$ or $[0, 1/(d^j+1)]$, respectively. This ensures that the actions lie within a $d^j$-dimensional unit cube (for the Beta distribution) or simplex (for the Dirichlet distribution), where these distribution families are applicable. 
For the \emph{Dirichlet} and \emph{Beta} baselines, since the normality score is not constant (c.f. Proposition \ref{th:z_expectation}), CUSUM is not applicable. Instead, we used the window-average-based method proposed by \cite{kia2023_decentralized_anomaly} to compute the decision metrics. For comparison, we also include results for replacing CUSUM with this metric in PGC in Appendix C.3.   
\color{black}
\subsection{Evaluation Methodology}

\begin{figure*}[t]
\centering
\subfigure[Multiwalker]{\includegraphics[width=0.23\textwidth]{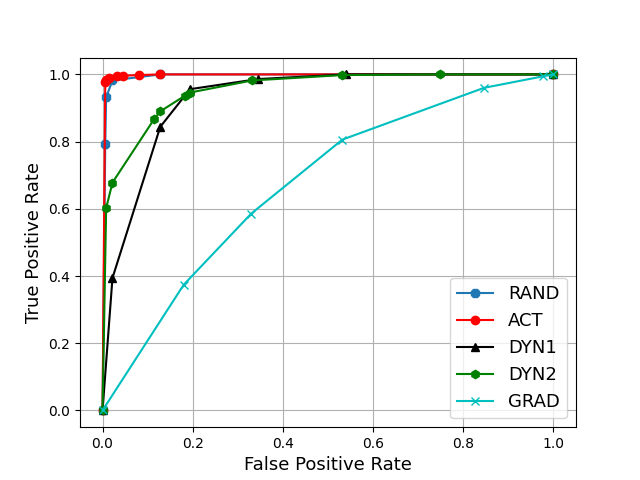}%
\label{fig:roc_Multiwalker}}
\hfil
\subfigure[Tag]{\includegraphics[width=0.23\textwidth]{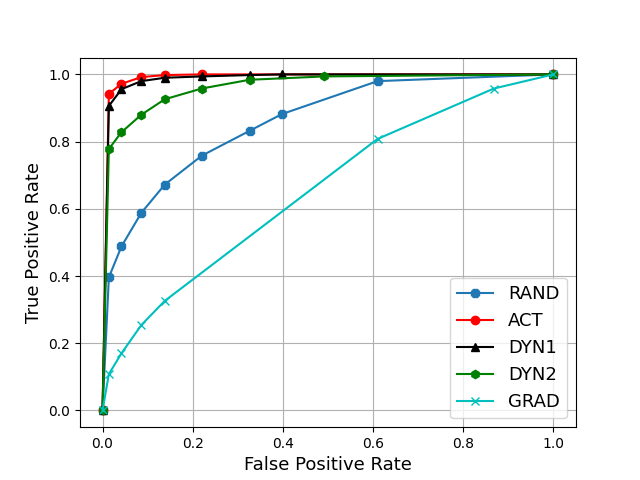}
\label{fig:roc_Tag}}
\hfil
\subfigure[World Comm]{\includegraphics[width=0.23\textwidth]{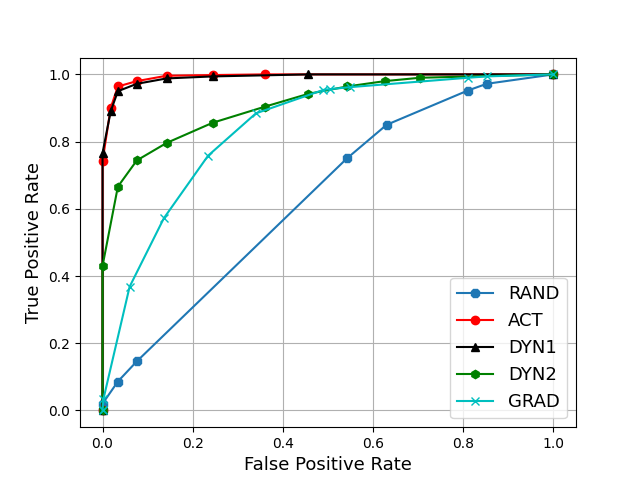}
\label{fig:roc_World_Comm}}
\subfigure[Pistonball]{\includegraphics[width=0.23\textwidth]{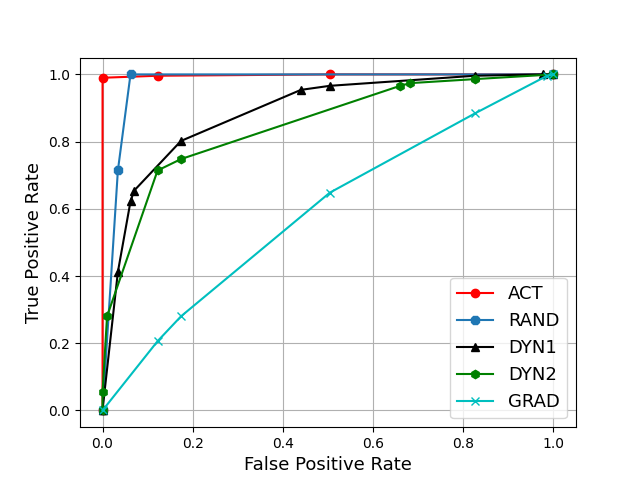}
\label{fig:roc_pistonball}}
\caption{ROC of the proposed method against different attacks in all environments.}
\label{fig:roc}
\end{figure*}
\begin{table*}[t]
    \caption{AUC score of PGC and Discrete against RAND, ACT, GRAD and DYN attacks in all environments. }
    
    \centering
    \begin{tabular}{c|c c| c c| c c| c c|}
    \hline
         & \multicolumn{2}{|c|}{\textbf{Multiwalker}} & \multicolumn{2}{|c|}{\textbf{Tag}} & \multicolumn{2}{|c|}{\textbf{World Comm}} & \multicolumn{2}{|c|}{\textbf{Pistonball}}\\ \hline
                   \textbf{Attack Types}     &  \emph{PGC}& \emph{Discrete}&  \emph{PGC}& \emph{Discrete}&  \emph{PGC}& \emph{Discrete}&  \emph{PGC}& \emph{Discrete}\\ \hline
         \textbf{ACT}    & \textbf{0.996}  &  0.972     &  \textbf{0.993} & 0.948      & \textbf{0.995} & 0.821     &  \textbf{0.999} & 0.758\\ \hline
         \textbf{RAND}   & \textbf{0.995}  &  0.855     &  0.843 &  \textbf{0.893}     & 0.677 & \textbf{0.713}      &  \textbf{0.997} &  0.970 \\ \hline
         \textbf{GRAD}   & \textbf{0.674}  &  0.566     &  0.653 &  \textbf{0.858}     & 0.884 & \textbf{0.913}      &  \textbf{0.581} & 0.554 \\ \hline
         \textbf{DYN1}   & \textbf{0.929}  &  0.818     &  \textbf{0.988} &  0.964     & \textbf{0.992} & 0.754      &  \textbf{0.907} & 0.711\\ \hline
         \textbf{DYN2}   & \textbf{0.954}  &  0.788     &  \textbf{0.968} & 0.944      & \textbf{0.912} & 0.707      &  \textbf{0.876} & 0.658\\ \hline
    \end{tabular}

     \label{tab:auc}
\end{table*}
Given the four environments and four attack strategies, we followed four steps for the evaluation.\footnote{Code available at https://github.com/kiarashkaz/PGC-Adversarial-Detection-Continuous}
\subsubsection{Step1: Training the team agents}
We used RLlib Python framework~\cite{rllib} to train the c-MARL agents. For Multiwalker and Pistonball, the multi-agent implementation of the PPO algorithm~\cite{ppo}, and for Tag and World Comm, the multi-agent APE-X DDPG algorithm \cite{horgan2018_apexDDPG} were used.  The selection of the algorithms was based on the results reported in \cite{terry2020_MARLalgs}. 

\subsubsection{Step2: Training $\text{NET}^{ij}$}
{\color{black}The considered structure for each $\text{NET}^{ij}$ is shown in Figure~\ref{fig:NET}. In Pistonball, we used the initial convolutional layers of the agents' policy networks, obtained during the c-MARL training stage, as the preprocessing network. In other environments, the preprocessing network consists of a fully connected layer followed by a ReLU activation, trained from scratch along with the other layers. We provide further details in Appendix D.} We trained $\text{NET}^{ij}$ over 5000 episodes for Multiwalker and Pistonball, and 20,000 episodes for Tag and World Comm. During the mentioned training phase the team agents used the policies learned in Step 1. 

\subsubsection{Step3: Training the Attacks}
We trained the ACT and the DYN attacks using a single-agent RL algorithm as explained in Section \ref{section_attacks}. We trained multiple DYN attacks  using different $\lambda$ values. For training each attack, we employed the single-agent PPO algorithm in RLlib over 20,000 episodes and selected the policy with the best adversarial reward.

\subsubsection{Step4: Evaluation of the Detector} 
For each attack (and also a scenario without any attack), we used the c-MARL polices, an attack policy, and a  detector for 500 episodes. \textbf{The detection rule} applied in our evaluations was as follows: if any of the non-victim agents triggered a detection, we considered the attack as detected at that particular time step (i.e., $u=1$). For the evaluation we considered the same set of detection thresholds for all agents. {\color{black}The value of hyperparameter $w$ is 0.5 in Multiwalker and World Comm, 0.75 in Tag and 0.05 in Pistonball. We provide results for other values of $w$ in Appendix C.1.} 

\color{black}

\subsection{Detection Performance}

\label{sec:roc_results}
 We first use the Receiver Operating Characteristic (ROC) curve to evaluate the performance of the proposed detector. The ROC curve shows the true positive rate against the false positive rate, generated by varying detection thresholds $\beta^{v^+}$ and $\beta^{v^-}$. The true positive rate is defined as the fraction of attacked episodes that are successfully detected, while the false positive rate is the fraction of unattacked episodes incorrectly classified as attacked. The detection performance can characterized using the Area Under the ROC Curve (AUC), where AUC=1 corresponds to a perfect detector. 
The ROC curves of all environments is shown in Figure \ref{fig:roc}.

\begin{table}[t]
    \caption{Average total episodic reward under different attacks.}
\centering
    \color{black}
    \small
    \begin{tabular}{m{0.18\linewidth}| m{0.17\linewidth} m{0.12\linewidth} m{0.12\linewidth} m{0.12\linewidth}}
         & {Multiwalker} & {Tag} & {World Comm} & {Pistonball}\\ \hline
        \textbf{No Attack}   & -12.7   &    101.8  &  37.6 & 228.6  \\ \hline
         \textbf{ACT}    & -107.6     &    64.9 &  26.7 & 83.1  \\ \hline
         \textbf{RAND}   & -75.6     &    68.1 &  30.4 & 202.1 \\ \hline
         \textbf{Grad}   & -42.7     &    90.4 &  34.7 & 215.1 \\ \hline
         \textbf{DYN1}   & -96.9     &    65.1 &  27.8 & 95.5 \\ \hline
         \textbf{DYN2}   & -89.6     &    69.2 &  30.1 & 139.3 \\ \hline
    \end{tabular}

    \label{tab:impact}
\end{table}
Figure \ref{fig:roc} and Table~\ref{tab:auc} show the ROC curves and the AUC scores of PGC against different attacks in all environments, respectively. DYN1 and DYN2 correspond to two DYN attacks trained with different values of $\lambda$, illustrating varying levels of detectability. 
To illustrate the corresponding attack impact, Table~\ref{tab:impact} shows the average total reward collected by the team agents under the different attacks. These results show that PGC can successfully detect the impactful attacks in all environments. 
The ACT attack, which causes the most significant reduction in team reward (e.g., from -12.7 to -107.6 in Multiwalker), can be detected almost perfectly, with an AUC score close to 1.
Detecting attacks with lower impact is more challenging. 
{\color{black} For example, GRAD in Pistonball or RAND in World Comm can partially bypass the detector (resulting in low AUC scores), but at the same time, the reduction in the team reward caused by these attacks is negligible.} One explanation for this observation is that the victim agent's original policy closely resembles these adversarial policies, making it unsurprising that distinguishing between the agent's normal behavior and these attacks is challenging. The trade-off between the attack impact and the detectability of the attack can be observed in other scenarios as well, which confirms that \textit{only those attacks can (partially) remain undetected that have very limited impact on the performance of the system.}  

\color{black}
Table~\ref{tab:auc} also shows that PGC outperforms \emph{Discrete} in most cases, especially against the more impactful attacks. Discrete performs better than PGC against GRAD in some scenarios. This can be explained by the fact that manipulated actions resulting from Grad are actually slight variations of the normal actions at each step. Thus they might not trigger a high anomaly score in PGC, but they might correspond to a different beam when distributions are modeled over a discrete set, resulting in a relatively higher anomaly score.

Another key aspect of comparison is complexity. \emph{PGC} has significantly lower computational cost than \emph{Discrete}. For example, in Tag, the detector networks for \emph{Discrete} require 243 outputs (one per action), while PGC requires only 20 outputs (one per action space dimension and one per covariance). This difference becomes even more pronounced in high-dimensional environments. For example, \emph{Discrete} requires neural networks with $3^9\approx 2\times10^4$ outputs, while this number is only 54 for PGC. The high output dimensionality of \emph{Discrete} necessitates significantly larger hidden layers for effective training, making it extremely computationally demanding to achieve performance comparable to that of PGC. 
\color{black}

Finally, an important metric for evaluating a detector’s performance is its time to detection. We provide the evaluation results for PGC in Appendix C.2.

\color{black}
\subsection{Why Multivariate Gaussian?}
\iffalse
\begin{figure}[t]
    \centering
    \includegraphics[width=0.8\linewidth]{figs/tag_IPGD.png}
    \caption{ROC of PGC and IPGC against the ACT and RAND attacks in \emph{Tag}.}
    \label{fig:ROC_IGPD}
\end{figure}
\fi

Recall that PGC uses a multivariate Gaussian distribution with a non-diagonal covariance matrix for the approximation of the action distribution. IPGC is a simpler alternative that assumes different dimensions of $a_t^j$ are independent, and thus, $\boldsymbol{\Sigma}_t^{ij}$ is diagonal. Hence, $\text{NET}^{ij}$ would have $2d$ outputs. While this simplification may not impact performance in certain scenarios, there are environments where the assumption of independence of action dimensions does not hold. 

Table~\ref{tab:auc_other_baselines} shows the AUC scores of PGC and IPGC against RAND in environments with an action dimension greater than 1. The results indicate significantly lower scores for IPGC, particularly in Tag and World Comm, where RAND is almost undetectable by IPGC, despite its impact in Tag. The poor performance of IPGC is due to the high correlation between elements of the action vector in non-compromised agents. Training independent predictor networks in such cases results in a predicted distribution with high variance, making the RAND attack undetectable. We note that, however, there might be environments where the elements of the action vector are close to independent, and in such environments IPGC can be used to reduce the complexity of detection.   

Table~\ref{tab:auc_other_baselines} shows the AUC scores obtained using Beta and Dirichlet distributions (which partially account for correlations between  dimensions) against RAND. The results shows that neither distribution produces an effective detector, as the RAND attack is nearly undetected in all scenarios. 

To further understand how accurate is the Gaussian approximation of the distributions $f^{ij}$, we compared the episodic average of the normality score under normal agent behavior with its expected value under the Gaussian assumption (Appendix A). We found that these quantities are very close in all environments, supporting that the Gaussian approximation may be accurate. Clearly, there may be MARL scenarios where $f^{ij}$ are multi-modal, in which case a centralized approach could be a more effective choice for detection. We provide a more detailed discussion in Appendix A. 
 
\begin{table}[t]
    \caption{AUC score of different detector baselines against RAND.}
    \centering
    \small
    \begin{tabular}{c|c c c}
        \textbf{Variation} & {Multiwalker} & {Tag} & {World Comm}\\ \hline
        \textbf{PGC}    & \textbf{0.99}     &    \textbf{0.84} &  \textbf{0.67}  \\ \hline
         \textbf{IPGC}    & 0.73     &    0.53 &  0.55  \\ \hline
         \textbf{Beta}   & 0.51     &    0.50 &  0.50  \\ \hline
         \textbf{Dirichlet}   & 0.54     &  0.50 &  0.50  \\ \hline
    \end{tabular}

     \label{tab:auc_other_baselines}
\end{table}
\color{black}
\subsection{Parameter Sharing for Reduced Complexity}
So far, we showed results when each agent has a separate detector for keeping track of every observable neighbor's action. This requires a total of $\sum_i|\mathcal{K}^i|$ predictor networks to be trained for detection,which can amount to as many as $K(K-1)$ networks in the worst case. In the following, we explore whether parameter sharing, a common approach in training MARL algorithms \cite{christianos2021_ParameterSharing}, \cite{terry2020_MARLalgs}, can be effectively applied in attack detection. With parameter sharing, a single predictor network is trained per potential victim agent, and the trained model is shared among all agents that observe that neighbor, reducing the total number of models to $K$. It is important to note that parameter sharing is applicable only to environments where the dimension of all agents' observations is the same, and there exists some form of symmetry among the observation spaces of the agents. 

\begin{table}[t]
    \caption{AUC score of PGC trained with parameter sharing against RAND and ACT.}
    \centering
    %\small
    \begin{tabular}{m{0.18\linewidth}| m{0.17\linewidth} m{0.12\linewidth} m{0.12\linewidth} m{0.12\linewidth}}
         & {Multiwalker} & {Tag} & {World Comm} & {Pistonball}\\ \hline
         \textbf{ACT}    & 0.995     &    0.994 &  0.997& 0.999 \\ \hline
         \textbf{RAND}   & 0.995     &    0.817 &  0.713& 0.997 \\ \hline      
    \end{tabular}

     \label{tab:auc_parameter_sharing}
\end{table}

Table \ref{tab:auc_parameter_sharing} shows the AUC score of PGC trained using parameter sharing. Comparing Table~\ref{tab:auc} and Table~\ref{tab:auc_parameter_sharing} shows that parameter sharing does not affect performance. Interestingly, in some cases, parameter sharing has even enhanced performance. This observation aligns with the findings presented by \cite{terry2020_MARLalgs}, which demonstrated that parameter sharing can enhance the training of MARL algorithms with continuous action spaces. The improvement is attributed to parameter sharing utilizing more trajectories to train a shared architecture, leading to potential improvements in sample efficiency.

\subsection{Multiple Victims}
\label{sec::multivictim}
Finally, we consider the case when there are more than one victim agents in the environment. Recall that each predictor is trained using the non-attacked behavior of individual agents. Consequently, the number of victim agents should not influence the training of the predictors. Furthermore, the computation of the normality score for an agent's actions is independent of the actions of other agents. This independence in normality scores for different agents should enable PGC to perform well in the presence of multiple victim agents. 

{\color{black}To illustrate this, we evaluated PGC against two victim agents in Multiwalker and Pistonball. We considered that detection occurs when both victims are detected. Table \ref{tab:two_victims} shows the team reward and the AUC scores obtained. Overall, the results confirm that PGC can be utilized even when there is more than one victim.} 
\begin{table}[t]
    \caption{Average total team reward and AUC scores of PGC when detecting ACT and RAND attacks against two victim agents.}
    \centering
    \small
    \begin{tabular}{c|c c| c c}
    &\multicolumn{2}{c|}{Multiwalker}& \multicolumn{2}{c}{Pistonball}\\ \hline
        \textbf{Attack Type} & {Reward} & {AUC} &  {Reward}& {AUC}\\ \hline
         \textbf{ACT}    & -109.6     &  0.998   &  75.8& 0.997 \\ \hline
         \textbf{RAND}   & -96.0     &   0.997  &  180.6& 0.983 \\ \hline      
    \end{tabular}

     \label{tab:two_victims}
\end{table}

\section{Conclusion}
\label{section_conc}
In this paper, we introduced a decentralized approach for detecting adversarial attacks on multi-agent RL systems with continuous action space. Our proposed scheme leverages the observations of the agents to predict the action distributions of other agents. We used RNNs to approximate these distributions as parameterized Gaussian densities. We put forward a low-complexity anomaly score, computed by comparing the predictions with the actual actions taken by the agents. {\color{black}We analytically showed that the proposed score has a constant mean and employed the CUSUM method to detect deviations from this value.} The proposed method was evaluated against various attack strategies with different levels of impact and detectability. Our results show that the proposed method outperformed its discrete counterpart, with only low-impact attacks having the potential to evade detection. Moreover, we empirically showed that the non-diagonal elements of the covariance matrix of the parameterized distributions used for detection are important for high detection performance. We also found that the computational complexity of training the detector could be reduced through parameter sharing without compromising the detection rate, and that the proposed detector is effective in scenarios with multiple victims.   

%%%%%%%%%%%%%%%%%%%%%%%%%%%%%%%%%%%%%%%%%%%%%%%%%%%%%%%%%%%%%%%%%%%%%%%%

%%% Use this environment to include acknowledgements (optional).
%%% This will be omitted in doubleblind mode.

\begin{ack}
This work was partly funded by the Swedish Research Council through projects 2020-03860 and 2024-05269, and by Digital Futures through the CLAIRE project. The computations were
enabled by resources provided by the National Academic Infrastructure for Supercomputing in Sweden (NAISS) at Linköping University partially funded by the Swedish Research Council through grant agreement no. 2022-06725.
\end{ack}

%%%%%%%%%%%%%%%%%%%%%%%%%%%%%%%%%%%%%%%%%%%%%%%%%%%%%%%%%%%%%%%%%%%%%%%%

%%% Use this command to include your bibliography file.

\bibliography{ref}
%%%%%%%%%%%%%%%%%%%%%%%%%%%%%%%%%%%%%%%%%%%%%%%%%%%%%%%%%%%%%%%
\newpage

\section*{Appendix}

\setcounter{section}{0}
\renewcommand\thesection{\Alph{section}}   
\setcounter{equation}{14}
\setcounter{table}{6}
\setcounter{figure}{4}

\section{Accuracy of the Gaussian Assumption}
Proposition 1 states that if $a^j_t|o_t^i$ is Gaussian then the normality score $z_t^{ij}$ has a known and constant expected value. To see why the Gaussian assumption may be reasonable, let us assume that the action of agent $j$ (policy) is determined by $o^j$ and $a^j|o^j\sim N(\mu(o^j), \Sigma)$. Then if $o^j|o^i$ is also Gaussian, then it can be shown that in some scenarios (e.g., when $\mu(o^j)$ is a linear function), $a^j|o^i$ is  Gaussian. 

To provide an empirical validation, we computed the episodic average of $z_t^{ij}$ across all environments and compared it with its expected value under the Gaussian assumption. Table \ref{tab:average_z} shows that the empirical average of the normality score under normal agent behavior closely matches its expected value under the Gaussian assumption. This indicates that the proposed Gaussian approximation is a reasonable choice for the considered environments. 

We acknowledge that there could be environments where the Gaussian assumption is a poor approximation.
For instance, if $o^j|o^i$ is multi-modal (e.g., a mixture model), then $a^j|o^i$ can also follow a multi-modal distribution and a single Gaussian may not approximate the resulting action distribution well. Notably, such a scenario implies that an agent’s observations are not informative about other agents' observations, which would hinder distributed anomaly detection in such environments. In such environments a centralized approach might be needed.

\begin{figure}[b]
    \centering
    \includegraphics[width=0.8\linewidth]{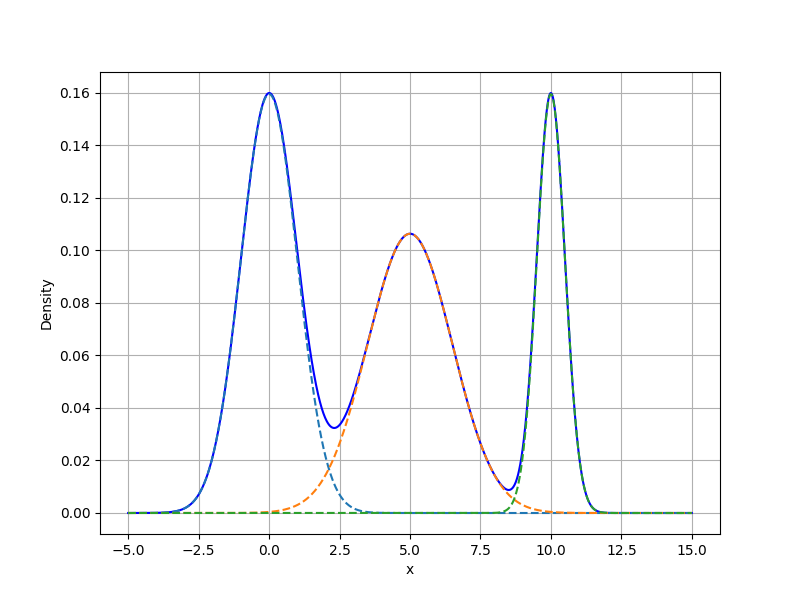}
    \caption{An example of one dimensional multi-modal distribution}
    \label{fig:Gaussian_mixture}
\end{figure}
\begin{table}[b]
    \centering
    \small
    \begin{tabular}{m{0.18\linewidth}| m{0.17\linewidth} m{0.12\linewidth} m{0.12\linewidth} m{0.12\linewidth}}
         & \small{Multiwalker} & \small{Tag} & \small{World Comm} & \small{Pistonball}\\ \hline
         $\mathbb{E}[z_t^{ij}]$ & -2  &  -2.5 & -4.5& -0.5 \\ \hline
         Average $z_t^{ij}$& -2.010  & -2.490 &  -4.441& -0.508 \\ \hline      
    \end{tabular}
    \caption{Episodic average and the expected value of normality score under the Gaussian assumption .}
     \label{tab:average_z}
\end{table}

\section{MARL Environments}
\begin{figure*}[t]
\centering
\subfigure[Multiwalker]{\includegraphics[width=0.23\textwidth]{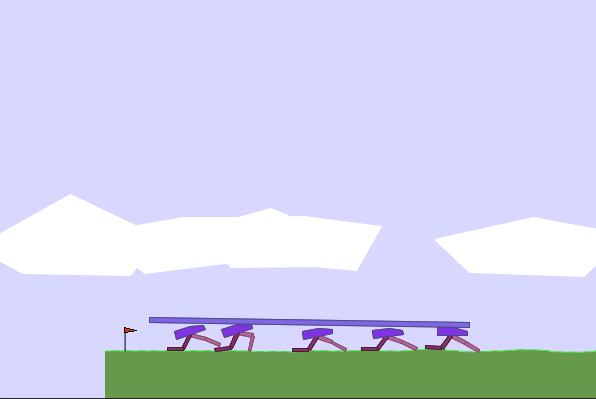}%
\label{fig:env_Multiwalker}}
\hfil
\subfigure[Tag]{\includegraphics[width=0.23\textwidth]{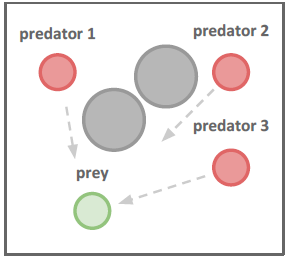}%
\label{fig:env_Tag}}
\hfil
\subfigure[World Comm]{\includegraphics[width=0.23\textwidth]{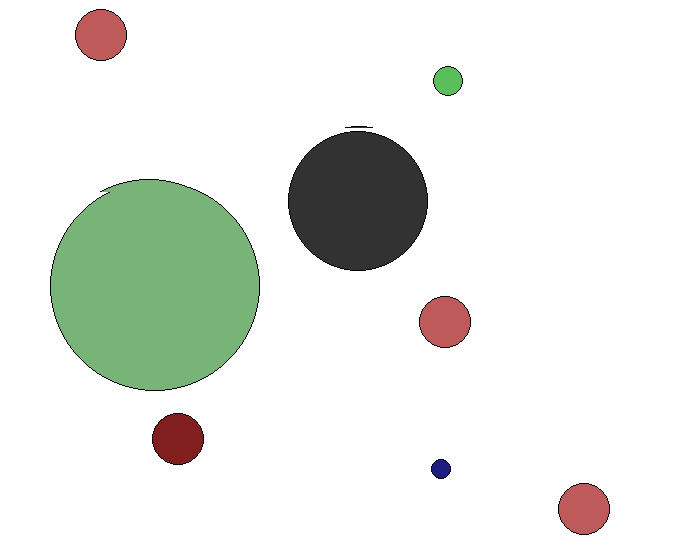}%
\label{fig:env_World_Comm}}
\subfigure[Pistonball]{\includegraphics[width=0.23\textwidth]{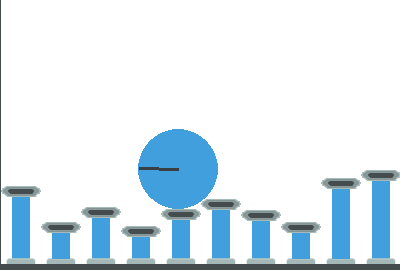}%
\label{fig:env_pistonball}}
\caption{Evaluated MARL Environments}
\label{fig:envs}
\end{figure*}
\begin{table*}[t]
    \centering
    \begin{tabular}{c|c c c| c c c| c c c|c c c|}
    \hline
         & \multicolumn{3}{|c|}{\textbf{Multiwalker}} & \multicolumn{3}{|c|}{\textbf{Tag}} & \multicolumn{3}{|c|}{\textbf{World Comm}} & \multicolumn{3}{|c|}{\textbf{Pistonball}}\\ \hline
                   \textbf{Attacks}     &  \footnotesize{$w=0$}&\footnotesize{$w=0.5$}&\footnotesize{$w=1$}&  \footnotesize{$w=0$}&\footnotesize{$w=0.75$}&\footnotesize{$w=1.5$}&  \footnotesize{$w=0$}&\footnotesize{$w=0.5$}&\footnotesize{$w=1$}&  \footnotesize{$w=0$}&\footnotesize{$w=0.05$}&\footnotesize{$w=0.4$}\\ \hline
         \textbf{ACT}    & 0.79  &  0.99 &  0.98 &        0.96  & 0.99 & 0.98      &      0.98& 0.99 & 0.99         &0.98 & 0.99 &0.97\\ \hline
         \textbf{RAND}   & 0.85  &  0.99 &  0.55 &        0.85 & 0.84 & 0.86       &      0.56&  0.67 & 0.63      & 0.99& 0.99&0.99 \\ \hline
         \textbf{GRAD}   & 0.57  &  0.67 &  0.68 &        0.63  & 0.65 & 0.64      &      0.81& 0.88  & 0.78       & 0.56& 0.58&0.52 \\ \hline
         \textbf{DYN1}   & 0.58  &  0.92 &  0.91 &         0.96 & 0.98 & 0.98      &      0.98& 0.99 & 0.98        & 0.72& 0.90& 0.8\\ \hline
         \textbf{DYN2}   & 0.70  &  0.95 &  0.91 &         0.94 & 0.96 & 0.96      &      0.91& 0.91 & 0.90        & 0.92 & 0.87& 0.67\\ \hline
    \end{tabular}
    \caption{AUC score of PGC obtained using different values of $w$.}
     \label{tab:w}
\end{table*}
\subsection{Multiwaker} 
In the \emph{Multiwalker}~\cite{gupta2017_SISL} environment, a set of bipedal robots cooperate to carry a package towards the right side of a terrain. The agents obtain a positive reward based on the change in the position of the package and a large negative reward in case the package falls. Each walker applies force to two joints, one in each leg, resulting in a 4-dimensional action space. The observation of each agent contains information about its legs and joints, as well as sensor measurements of the agent vicinity and neighboring agents. This environment is particularly interesting for our problem, due to the noisy nature of the sensor measurements. We considered a scenario with 5 agents for this benchmark.

\subsection{Tag}
\emph{Tag} (Simple Tag)~\cite{lowe2017_MPE} is one of the Multi-agent Particle Environments (MPE) family. In this environment, a team of three predators try to hunt a single prey. Predators are rewarded based on the number of times they can hit the prey in a limited duration (25 time-steps). Each agent's action is a 5-dimensional vector determining how the agent moves. Each agent observes the other agents' velocities and relative positions as part of the observation vector. 

\subsection{World Comm}
\emph{World Comm}~\cite{lowe2017_MPE} is another MPE scenario including two competing groups of agents, which we refer to as red and green agents. Green agents' objective is to collect food items and escape from red agents. Red agents are rewarded based on the proximity to green agents. There are forests where agents can hide themselves. We considered a scenario with 4 red agents and one green agent. The red team is trained using c-MARL and the green agent behaves based on a predefined policy. One of the red agents is the leader and is capable of observing all agents at all times and can coordinate its team using a 9-dimensional action vector. Other agents' actions are 5-dimensional.

\subsection{Pistonball}
\emph{Pistonball} is one of the Butterfly environments in PettingZoo, built on visual Atari spaces and powered by the Chipmunk physics engine. In this environment, a group of pistons collaborates to move a ball from the right side to the left side of the game border using vertical movements. Each piston's action is a scalar in (-1, 1), representing the amplitude and direction of its vertical movement. The observation for each piston is an RGB image of its surrounding area, including neighboring agents. The reward is determined by the extent of the ball's movement toward the left side. We considered a scenario with 10 agents.

\section{Other Results}

\subsection{Effect of hyperparameter $w$}
A hyperparameter for CUSUM detection in (6) is $w$, which controls the sensitivity of the detection. Table \ref{tab:w} shows the AUC score of PGC with different values of $w$.

%\subsection{ROC Curves in All Environments}

\begin{figure*}[t]
\centering
\subfigure[Multiwalker]{\includegraphics[width=0.23\textwidth]{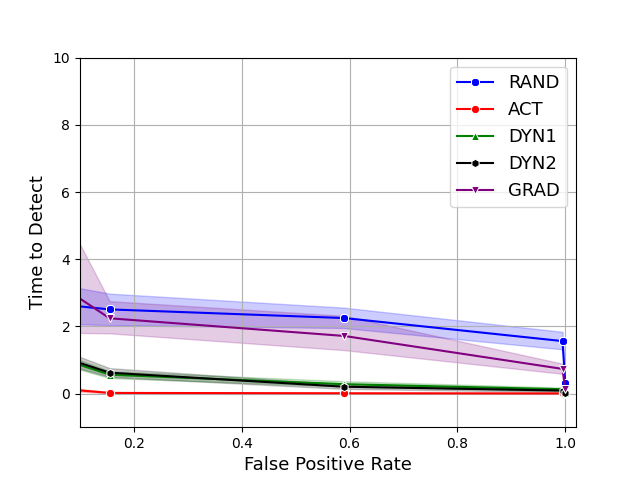}%
\label{fig:TTD_Multiwalker}}
\hfil
\subfigure[Tag]{\includegraphics[width=0.23\textwidth]{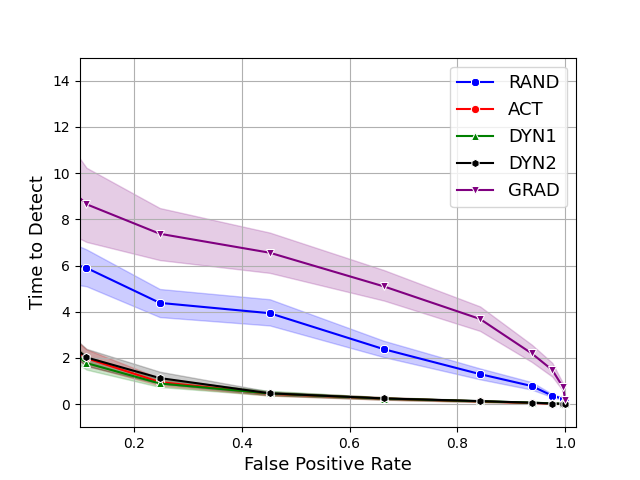}%
\label{fig:TTD_Tag}}
\hfil
\subfigure[World Comm]{\includegraphics[width=0.23\textwidth]{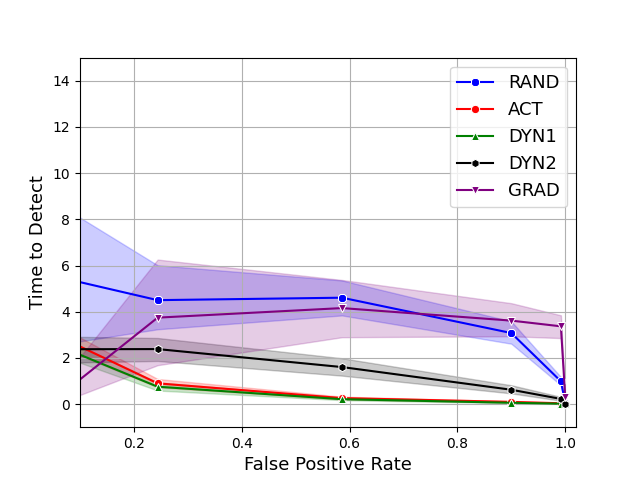}%
\label{fig:TTD_World_Comm}}
\hfil
\subfigure[Pistonball]{\includegraphics[width=0.23\textwidth]{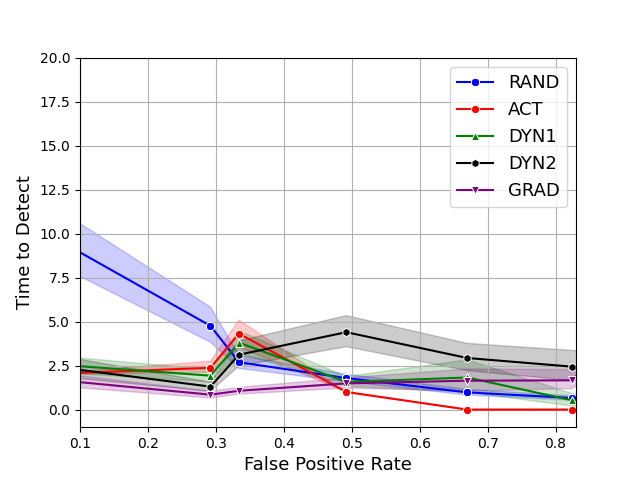}%
\label{fig:TTD_pistonball}}
\vspace{0.3in}
\caption{Time to detect vs. false positive rate for the proposed method against different attacks}
\label{fig:TTD}
\end{figure*}
\subsection{Time to Detection}
One important metric for the performance assessment of a detector is the time to detection. We define this metric as the average number of time steps from the start of the attack until it is detected. The averaging is taken only over true positive episodes, i.e., episodes in which a detection happens are considered. 
Figure \ref{fig:TTD} shows the time to detect as a function of false positive rate in all environments. The figure shows that all the impactful attacks (e.g., ACT) can be detected by PGC in less than 5 time steps (or even instantaneously in Multiwalker) already at a very low false positive rate. 
Another notable observation is that the time to detect of impactful attacks is not significantly affected by their stealthiness, as it takes almost the same time to detect the ACT and the DYN attacks in most of the scenarios.

\subsection{Window-Based Detection}
A variation of our proposed detector replaces CUSUM with a window-based averaging of the normality score. Table~\ref{tab:window_auc} presents the AUC scores of this scheme against various attacks across all environments. Comparing these results with the AUC scores obtained by PGC (Table 2) reveals that CUSUM significantly enhances detection performance overall.
\begin{table}[b]
\centering
    \color{black}
    \small
    \begin{tabular}{m{0.18\linewidth}| m{0.17\linewidth} m{0.12\linewidth} m{0.12\linewidth} m{0.12\linewidth}}
         & \small{Multiwalker} & \small{Tag} & \small{World Comm} & \small{Pistonball}\\ \hline
         \textbf{ACT}    & 0.99     &   0.97 &  0.97 & 0.99  \\ \hline
         \textbf{RAND}   & 0.68     &    0.80 &  0.55 & 0.83 \\ \hline
         \textbf{Grad}   & 0.70     &    0.57 &  0.87 & 0.61 \\ \hline
         \textbf{DYN1}   & 0.96     &    0.97 &  0.97 & 0.90 \\ \hline
         \textbf{DYN2}   & 0.89     &    0.95 &  0.97 & 0.81 \\ \hline
    \end{tabular}
    \caption{AUC score of the proposed detector with window-based detection.}
    \label{tab:window_auc}
\end{table}

\subsection{Observation Tracking Baseline}
As another baseline for comparison, we evaluated an anomaly detection approach based on the sequence of observations. Specifically, we adapted the detection method proposed by \cite{Zhang2024tmlr} to the multi-agent setting, where each agent independently tracks its own observation sequence to detect potential anomalies. This approach is justified by the fact that the victim agent's actions can cause deviations in the observation trajectories of non-victim agents, making it possible for them to detect anomalies affecting the c-MARL team. However, unlike our method, this approach cannot pinpoint the exact source of these abnormalities, i.e., the victim agent.

In this method, which we refer to as \emph{Observation Tracking}, for each action type, the mapped feature vectors of observations leading to that action are fitted to a multivariate Gaussian distribution. During execution, the minimum Mahalanobis distance between a new observation and any of the Gaussian models is used as the normality score. If this score exceeds a predefined threshold, an anomaly is detected. Since this method is designed for discrete action setups, we applied the same quantization technique as in \emph{Discrete} with $N_q=3$. For feature selection, we followed the approach in \cite{Zhang2024tmlr}, using 30 components for PCA transformation. 

Table \ref{tab:obs_tracking_auc} presents the AUC scores of the Observation Tracking method for detecting various attacks in Multiwalker and Tag. The results highlight the limitations of this approach in distributed detection of attacks on agents' actions, emphasizing the necessity of directly tracking actions rather than relying solely on observation sequences. 
\begin{table}[t]
\centering
    \color{black}
    \small
    \begin{tabular}{m{0.18\linewidth}| m{0.17\linewidth} m{0.12\linewidth} }
         & \small{Multiwalker} & \small{Tag} \\ \hline
         \textbf{ACT}    & 0.61     &   0.55  \\ \hline
         \textbf{RAND}   & 0.5     &    0.5 \\ \hline
         \textbf{Grad}   & 0.5     &    0.5  \\ \hline
         \textbf{DYN1}   & 0.55     &    0.52  \\ \hline
         \textbf{DYN2}   & 0.52     &    0.51  \\ \hline
    \end{tabular}
    \caption{AUC score of the observation tracking approach.}
    \label{tab:obs_tracking_auc}
\end{table}
\section{Hyperparameters} 
\subsection{Training Detectors and Attacks}
The hyperparameters for training detectors are shown in table \ref{tab:detectors_params}. The value of $\lambda$ used for training the DYN1 attack in Multiwalker, Tag, World Comm, and Pistonball, were 5, 1, 0.5, and 1, respectively. For DYN2 attack these values were 10, 2, 5, and 2, respectively.
 \begin{table}[t]
    \centering
    \begin{tabular}{c c}
         Hyperparameter & Value \\ \hline \hline
          Use\textunderscore rnn     &    True   \\ 
         hidden\textunderscore size (Pistonball)    &    256   \\ 
         hidden\textunderscore size (other environments)    &    128   \\ 
             batch\textunderscore size    &    20   \\ 
    include\textunderscore last\textunderscore act(Multiwalker)         &True   \\  
    include\textunderscore last\textunderscore act(Tag/World Comm)      &    False   \\ 
        lr     &    5e-4   \\ \hline                                 
    \end{tabular}
    \caption{Hyperparameters used for training predictors.}
     \label{tab:detectors_params}
\end{table}

 \subsection{Training c-MARL}
 The parameters we used for training c-MARL policies are summarized in Table \ref{tab:cmarl_params}. 
 
 \begin{table}[t]
 \small
    \centering
    \begin{tabular}{c|c c c}
        \textbf{Environment} &  Hyperparameter & Value \\ \hline
         \textbf{Multiwalker}    & Algorithm     &    PPO   \\ 
                                 & gamma     &    0.99   \\ 
                                 & kl\textunderscore coeff     &    0.2   \\ 
                                 & kl\textunderscore target     &    0.01   \\  
                                 & lambda     &    1   \\ 
                                 & lr     &    5e-5   \\
                                  & use\textunderscore lstm     &    True   \\
                                  & lstm\textunderscore cell\textunderscore size     &    128   \\
                                  & max\textunderscore seq\textunderscore len     &    25   \\
                                  & post\textunderscore fcnet\textunderscore activation     &    relu   \\
                                  &  vf\textunderscore share\textunderscore layers    &    False   \\
                                  & sgd\textunderscore minibatch\textunderscore size     &    128   \\ \hline
         \textbf{Tag/World Comm}  &  Algorithm     &    APE-X DDPG   \\
                                  & gamma     &    0.99   \\ 
                                  &  clip\textunderscore actions &  False \\
                                 & actor\textunderscore hiddens     &    [400, 300]   \\   
                                 & critic\textunderscore hiddens     &    [400, 300]   \\ 
                                 & lr     &    0.0005   \\
                                 & adam\textunderscore epsilon   & 1e-8 \\
                                  & train\textunderscore batch\textunderscore size     &    512   \\
                                  & use\textunderscore lstm &    False   \\
                                  & vf\textunderscore share\textunderscore layers     &    True   \\
                                  & compress\textunderscore observations  &  False \\
                                  & post\textunderscore fcnet\textunderscore activation     &    relu   \\ \hline
         \textbf{Pistonball}    & Algorithm     &    PPO   \\ 
                                 & gamma     &    0.99   \\ 
                                 & kl\textunderscore coeff     &    0.2   \\ 
                                 & kl\textunderscore target     &    0.01   \\  
                                 & lambda     &    1   \\ 
                                 & lr     &    5e-5   \\
                                  & use\textunderscore lstm     &    False   \\
                                  & post\textunderscore fcnet\textunderscore activation     &    relu   \\
                                  &  vf\textunderscore share\textunderscore layers    &    False   \\
                                  & sgd\textunderscore minibatch\textunderscore size     &    128   \\ \hline

    \end{tabular}
    \caption{Hyperparameters used for training c-MARL.}
     \label{tab:cmarl_params}
\end{table}

%\subsection{Effect of Hyperparameter $w$ on the Detection Performance}
%\vspace{-1em}
%\begin{figure}[h!]
%    \centering
%    \includegraphics[width=0.7\linewidth]{multiwalker_window.png}
%    \caption{AUC score as a function of window size ($w$) in \emph{Multiwalker}.}
%    \label{fig:window}
%\end{figure}

\end{document}